\documentclass[journal,a4paper,twoside]{sty/IEEEtran}

\usepackage [cp1250] {inputenc}
\usepackage{sty/EVjour}

\ifCLASSINFOpdf
   \usepackage[pdftex]{graphicx}
\else 
   \usepackage[dvips]{graphicx}
\fi
\usepackage{epstopdf} 

\usepackage{textcomp} 

\usepackage{amsfonts}
\usepackage{amssymb}
\usepackage{latexsym}
\usepackage{graphicx}
\usepackage{lscape}
\usepackage{multirow}
\usepackage{rotating}
\usepackage{amsmath}
\usepackage{color}

\begin{document}

\title{A Brief Review of Nature-Inspired Algorithms for Optimization}

\authors{ Iztok Fister Jr.${}^{1}$, Xin-She Yang${}^{2}$, Iztok Fister${}^{1}$, Janez Brest${}^{1}$, Du\v{s}an Fister${}^{1}$}

\address{$^1$University of Maribor
\\ Faculty of electrical engineering and computer science, Smetanova 17, 2000 Maribor, Slovenia
\\ E-mail: iztok.fister@guest.arnes.si
\\$^2$University of Middlesex
\\ School of Science and Technology, Middlesex University, London NW4 4BT, United Kingdom
\\ E-mail: xy227@cam.ac.uk}

\abstract{Swarm intelligence and bio-inspired algorithms form a hot topic in the developments of new algorithms inspired by nature. These nature-inspired metaheuristic algorithms can be based on swarm intelligence, biological systems, physical and chemical systems. Therefore, these algorithms can be called swarm-intelligence-based, bio-inspired, physics-based and chemistry-based, depending on the sources of inspiration. Though not all of them are efficient, a few algorithms have proved to be very efficient and thus have become popular tools for solving real-world problems. Some algorithms are insufficiently studied.
The purpose of this review is to present a relatively comprehensive list of all the algorithms in the literature, so as to inspire further research.}


\review{15 July 2013}    

\markboth{Fister, Yang Fister, Brest, Fister}{A Brief Review of Nature-Inspired Optimization Algorithms}

\maketitle

\IEEEpeerreviewmaketitle

\vspace{2cm}

\section{Introduction}

Real-world optimization problems are often very challenging to solve, and many applications have to deal with NP-hard problems. To solve such problems,
optimization tools have to be used, though there is no guarantee that the optimal solution can be obtained. In fact, for NP-problems, there are no
efficient algorithms at all. As a result, many problems have to be solved by trial and errors using various optimization techniques.  In addition, new algorithms have been developed to see if they can cope with these challenging optimization problems.

Among these new algorithms, many algorithms such as particle swarm optimization, cuckoo search and firefly algorithm,
have gained popularity due to their high efficiency. In the current literature, there are about 40 different algorithms.
It is really a challenging task to classify these algorithms systematically. Obviously, the classifications can largely
depend on the criteria, and there is no easy guideline to set out the criteria in the literature.
As criteria may vary, detailed classifications can be an impossible task for a research paper.
However, in this short paper, we only attempt to focus on one aspect of the characteristics of
these algorithms. That is, we will focus on the source of inspiration when developing algorithms.

Therefore, the rest of this paper is organized as follows: Section 2 analyzes the sources
of inspiration, while Section 3 provides a brief and yet comprehensive list
of algorithms. Finally, Section 4 concludes with some suggestions.

\section{Sources of Inspiration}

Nature has inspired many researchers in many ways and thus is a rich source of inspiration.
Nowadays, most new algorithms are nature-inspired, because they have been developed
by drawing inspiration from nature. Even with the emphasis on the source of inspiration,
we can still have different levels of
classifications, depending on how details and how many subsources we will wish to use.
For simplicity, we will use the highest level sources such as biology, physics or chemistry.

In the most generic term, the main source of inspiration is Nature. Therefore, almost all
new algorithms can be referred to as nature-inspired. By far the majority of nature-inspired
algorithms are based on some successful characteristics of biological system. Therefore, the
largest fraction of nature-inspired algorithms are biology-inspired, or bio-inspired for short.

Among bio-inspired algorithms, a special class of algorithms have been developed by drawing
inspiration from swarm intelligence. Therefore, some of the bio-inspired algorithms can be
called swarm-intelligence-based. In fact, algorithms based on swarm intelligence are among the
most popular. Good examples are ant colony optimization \cite{dorigo1992optimization},
particle swarm optimization \cite{kennedy1995particle},
cuckoo search \cite{yang2009cuckoo},  bat algorithm \cite{Bat:2010}, and firefly algorithm \cite{yang2008book,fister2013comprehensive}.

Obviously, not all algorithms were based on biological systems. Many algorithms have been
developed by using inspiration from physical and chemical systems. Some may even be based on
music \cite{geem2001new}. In the rest of paper, we will briefly divide all algorithms into
different categories, and we do not claim that this categorization is unique. This is a good attempt
to provide sufficiently detailed references.

\section{Classification of Algorithms}

Based on the above discussions, we can divide all existing algorithms into four major
categories: swarm intelligence (SI) based, bio-inspired (but not SI-based), physics/chemistry-based,
and others. We will summarize them briefly in the rest of this paper.
However, we will focus here on the relatively new algorithms. Well-established algorithms such as genetic algorithms
are so well known that there is no need to introduce them in this brief paper.

It is worth pointing out the classifications here are not unique as some algorithms can be classified into
different categories at the same time. Loosely speaking, classifications depend largely on
what the focus or emphasis and the perspective may be. For example, if the focus and perspective
are about the trajectory of the search path, algorithms can be classified as trajectory-based
and population-based. Simulated annealing is a good example of trajectory-based algorithms,
while particle swarm optimization and firefly algorithms are population-based algorithms.
If our emphasis is placed on the interaction of the multiple agents, algorithms can be
classified as attraction-based or non-attraction-based. Firefly algorithm (FA) is a good example
of attraction-based algorithms because FA uses the attraction of light and attractiveness
of fireflies, while genetic algorithms are non-attraction-based since there is no explicit
attraction used. On the other hand, if the emphasis is placed on the updating equations,
algorithms can be divided into rule-based and equation-based. For example, particle swarm optimization
and cuckoo search are equation-based algorithms because both use explicit updating equations,
while genetic algorithms do not have explicit equations for crossover and mutation.
However, in this case, the classifications are not unique. For example, firefly algorithm
uses three explicit rules and these three rules can be converted explicitly into a single
updating equation which is nonlinear \cite{yang2008book,fister2013comprehensive}. This clearly
shows that classifications depend on the actual perspective and motivations.
Therefore, the classifications here are just
one possible attempt, though the emphasis is placed on the sources of inspiration.

\subsection{Swarm intelligence based}

Swarm intelligence (SI) concerns the collective, emerging behaviour of multiple, interacting agents
who follow some simple rules. While each agent may be considered as unintelligent, the
whole system of multiple agents may show some self-organization behaviour and thus can behave like
some sort of collective intelligence. Many algorithms have been developed by drawing inspiration
from swarm-intelligence systems in nature.

All SI-based algorithms use multi-agents, inspired by the collective behaviour of social insects, like ants, termites, bees, and wasps, as well as from other animal societies like flocks of birds or fish. A list of swarm intelligence algorithms is presented in Table~\ref{Table1}.
The classical particle swarm optimization (PSO) uses the swarming behaviour of fish and birds,
while firefly algorithm (FA) uses the flashing behaviour of swarming fireflies.
Cuckoo search (CS) is based on the brooding parasitism of some cuckoo species,
while bat algorithm uses the echolocation of foraging bats. Ant colony optimization
uses the interaction of social insects (e.g., ants), while the class of bee algorithms are all
based on the foraging behaviour of honey bees.

SI-based algorithms are among the most popular and widely used. There are many reasons for such popularity, one of the reasons is that SI-based algorithms usually sharing information among multiple agents, so that self-organization, co-evolution and learning during iterations may help to provide the high efficiency of most SI-based algorithms. Another reason is that multiple agent can be parallelized easily so that large-scale optimization becomes more practical
from the implementation point of view.

\subsection{Bio-inspired, but not SI based}

Obviously, SI-based algorithms belong to a wider class of algorithms, called bio-inspired algorithms.
In fact, bio-inspired algorithms form a majority of all nature-inspired algorithms.
From the set theory point of view, SI-based algorithms are a subset of bio-inspired algorithms,
while bio-inspired algorithms are a subset of nature-inspired algorithms.
That is
\[ \textrm{SI-based} \subset \textrm{bio-inspired} \subset \textrm{nature-inspired}. \]
Conversely, not all nature-inspired algorithms are bio-inspired, and some are purely physics and chemistry based algorithms
as we will see below.

Many bio-inspired algorithms do not use directly the swarming behaviour.
Therefore, it is better to call them bio-inspired, but not SI-based.
For example, genetic algorithms are bio-inspired, but
not SI-based. However, it is not easy to classify certain algorithms such as differential evolution (DE). Strictly speaking,
DE is not bio-inspired because there is no direct link to any biological behaviour. However, as it has some similarity
to genetic algorithms and also has a key word `evolution', we tentatively put it in the category of bio-inspired algorithms.
These relevant algorithms are listed in Table~\ref{Table1}.

For example, the flower algorithm \cite{yang2012flower}, or flower pollination algorithm \cite{yangflower2013}, 
developed by Xin-She Yang in 2012 is a bio-inspired algorithm, but it is not a SI-based algorithm because
flower algorithm tries to mimic the pollination characteristics of flowering plants and the associated
flower consistency of some pollinating insects.

\subsection{Physics and Chemistry Based}

Not all metaheuristic algorithms are bio-inspired, because their sources of inspiration 
often come from physics and chemistry.
For the algorithms that are not bio-inspired, most have been developed by mimicking certain
physical and/or chemical laws, including electrical charges, gravity, river systems, etc.
As different natural systems are relevant to this category, we can even subdivide
these into many subcategories which is not necessary.
A list of these algorithms is given in Table~\ref{Table1}.

Schematically, we can represent the relationship of physics and chemistry based algorithms
as the follows:
\[
\frac{\textrm{Physics algorithms}}{\text{Chemistry algorithms}} \left\{ \begin{array}{l}
 \notin  \textrm{bio-inspired algorithms} \\ \\
 \in  \text{nature-inspired algorithms}
\end{array} \right.
\]
Though physics and chemistry are two different subjects, however, it is not useful to
subdivide this subcategory further into physics-based and chemistry. After all, many
fundamental laws are the same. So we simply group them as physics and chemistry based algorithms.

\subsection{Other algorithms}
When researchers develop new algorithms, some may look for inspiration away from nature.
Consequently, some algorithms are not bio-inspired or physics/chemistry-based,  it 
is sometimes difficult to put some algorithms in the above three categories, because these
algorithms have been developed by using various characteristics from different sources, such as
social, emotional, etc. In this case, it is better to put them in the other category, as listed
in Table~\ref{Table1}

\subsection{Some Remarks}
Though the sources of inspiration are very diverse, the algorithm designed from such
inspiration may be equally diverse. However, care should be taken, as true novelty is
a rare thing. For example, there are about 28,000 living species of fish,
this cannot mean that researchers should develop 28000 different algorithms based on fish.
Therefore, one cannot call their algorithms trout algorithm, squid algorithm, ..., shark algorithm.

In essence, researchers try to look for some efficient formulas as summarized by Yang
\cite{yangswarmbook2013} as the following generic scheme:
\[ [x_1, x_2, ..., x_n]^{t+1}=A\big\{[x_1, x_2, ..., x_n]^t; ...; \]
\[ \qquad (p_1,p_2, ..., p_k); (w_1, w_2, ..., w_m)  \big\}, \]
which attempts to generate better solutions (a population of $n$ solutions) at iteration $t+1$ from
the current iteration $t$ and its solution set $x_i, (i=1,2,...,n)$. This iterative algorithmic engine
(i.e., algorithm $A$) also uses some algorithm-dependent parameters $(p_1, ..., p_k)$
and some random variables $(w_1, ..., w_m)$. This schematic representation can include
all the algorithms listed in this paper. However, this does not mean it is easy
to analyze the behaviour of an algorithm because this formula can be highly nonlinear.
Though Markov chains theory and dynamical system theory can help to
provide some limited insight into some algorithms, the detailed mathematical framework
is still yet to be developed. 

On the other hand, it is worth pointing out that studies show that some algorithms are better
than others. It is still not quite understood why. However, if one looks at the intrinsic part
of algorithm design closely, some algorithms are badly designed, which lack certain basic capabilities
such as the mixing and diversity among the solutions. In contrast, good algorithms have
both mixing and diversity control so that the algorithm can explore the vast search
space efficiently, while converge relatively quickly when necessary.
Good algorithms such as particle swarm optimization, differential evolution, cuckoo search 
and firefly algorithms all have both global search and intensive local search capabilities,
which may be partly why they are so efficient.

\section{Conclusion}
The sources of inspiration for algorithm development are very diverse, and consequently the algorithms are equally diverse.
In this paper, we have briefly summarized all the algorithms into 4 categories. This can be a comprehensive
source of information to form a basis or starting point for further research. It is worth pointing out that
the classifications may not be unique, and this present table is just for the purpose of information only.

Based on many studies in the literature, some algorithms are more efficient and popular than others. It would be helpful
to carry out more studies, but this does not mean that
we should encourage researchers  to develop more algorithms such as grass, sky, or ocean algorithms.

Currently, there may be some confusion and distraction in the research
of metaheuristic algorithms. On the one hand, researchers have focused
on important novel ideas for solving difficult problems. On the other hand,
some researchers artificially invent new algorithms for the sake of
publications with little improvement and no novelty.
Researchers should be encouraged to carry out truly novel and important
studies that are really useful to solve hard problems. Therefore, our
aim is to inspire more research to gain better insight into efficient
algorithms and solve large-scale real-world problems.

\begin{landscape}
\begin{table}
\begin{tabular}{ |l|l|l|l|l|l|l|l|l| }
\hline
\multicolumn{3}{ |c| }{\textcolor{blue}{Swarm intelligence based algorithms}} & \multicolumn{3}{ |c| }{\textcolor{blue}{Bio-inspired (not SI-based) algorithms}}\\
\hline
 Algorithm & Author & Reference & Algorithm & Author & Reference \\ \hline
   Accelerated PSO & Yang et al. & \cite{yang2008book,yang2010natureluniver} & Atmosphere clouds model & Yan and Hao & \cite{yan2013novel} \\
   Ant colony optimization & Dorigo& \cite{dorigo1992optimization} & Biogeography-based optimization & Simon & \cite{simon2008biogeography} \\
   Artificial bee colony & Karaboga and Basturk & \cite{karaboga2007powerful} & Brain Storm Optimization & Shi & \cite{shi2011optimization}\\
   Bacterial foraging & Passino & \cite{passino2002biomimicry} & Differential evolution & Storn and Price & \cite{storn1997differential}\\
   Bacterial-GA Foraging & Chen et al. & \cite{chen2007novel} & Dolphin echolocation & Kaveh and Farhoudi & \cite{kaveh2013new}\\
   Bat algorithm & Yang & \cite{Bat:2010} & Japanese tree frogs calling & Hern{\'a}ndez and Blum & \cite{hernandez2012distributed}\\     	
   Bee colony optimization & Teodorovi{\'c} and Dell'Orco & \cite{teodorovic2005bee} & Eco-inspired evolutionary algorithm & Parpinelli and Lopes & \cite{parpinelli2011eco}\\
   Bee system & Lucic and Teodorovic & \cite{lucic2001bee} & Egyptian Vulture & Sur et al. & \cite{sur2013egyptian}\\
   BeeHive & Wedde et al. & \cite{Wedde200483} & Fish-school Search & Lima et al. & \cite{de2008novel,bastos2009fish}\\
    Wolf search & Tang et al. & \cite{6360147} & Flower pollination algorithm & Yang & \cite{yang2012flower,yangflower2013}\\
   Bees algorithms & Pham et al. & \cite{pham2006bees} & Gene expression & Ferreira & \cite{ferreira2001gene}\\
   Bees swarm optimization & Drias et al. & \cite{drias2005cooperative} & Great salmon run & Mozaffari & \cite{mozaffari2012great}\\
   Bumblebees & Comellas and Martinez & \cite{comellas2011bumblebees} & Group search optimizer & He et al. & \cite{he2009group}\\
   Cat swarm & Chu et al. & \cite{Chu2006854} & Human-Inspired Algorithm & Zhang et al. & \cite{zhang2009human} \\
   Consultant-guided search & Iordache & \cite{iordache2010consultant} & Invasive weed optimization & Mehrabian and Lucas & \cite{mehrabian2006novel}\\
   Cuckoo search & Yang and Deb & \cite{yang2009cuckoo} & Marriage in honey bees & Abbass & \cite{abbass2001mbo}\\
   Eagle strategy & Yang and Deb & \cite{yang2010eagle} & OptBees & Maia et al. & \cite{maia2012bee}\\
   Fast bacterial swarming algorithm & Chu et al. & \cite{chu2008fast} & Paddy Field Algorithm & Premaratne et al. & \cite{premaratne2009new}\\
   Firefly algorithm & Yang & \cite{yang2010firefly} & Roach infestation algorithm& Havens & \cite{havens2008roach}\\
   Fish swarm/school & Li et al. & \cite{Li200232} & Queen-bee evolution & Jung & \cite{jung2003queen}\\
   Good lattice swarm optimization & Su et al. & \cite{su2007good}  & Shuffled frog leaping algorithm & Eusuff and Lansey & \cite{Eusuff2003210}\\
   Glowworm swarm optimization & Krishnanand and Ghose & \cite{krishnanand2005detection,krishnanand2009glowworm} & Termite colony optimization & Hedayatzadeh et al. & \cite{hedayatzadeh2010termite} \\
   Hierarchical swarm model & Chen et al. & \cite{chen2010hierarchical} & \multicolumn{3}{ |c| }{\textcolor{blue}{Physics and Chemistry based algorithms}}  \\
   Krill Herd & Gandomi and Alavi & \cite{gandomi2012krill} & Big bang-big Crunch & Zandi et al. & \cite{zandi2012reactive}\\
    Monkey search & Mucherino and Seref & \cite{mucherino2007monkey} &  Black hole & Hatamlou & \cite{hatamlou2012black}\\
   Particle swarm algorithm & Kennedy and Eberhart & \cite{kennedy1995particle} & Central force optimization & Formato & \cite{formato2007central}\\
   Virtual ant algorithm & Yang & \cite{yang2006application} & Charged system search & Kaveh and Talatahari & \cite{kaveh2010novel}\\
   Virtual bees & Yang & \cite{Yang2005317} & Electro-magnetism optimization & Cuevas et al. & \cite{cuevas2012circle}\\
  Weightless Swarm Algorithm & Ting et al. & \cite{ting2012weightless} & Galaxy-based search algorithm & Shah-Hosseini & \cite{shah2011principal}\\
   \multicolumn{3}{ |c| }{\textcolor{blue}{Other algorithms}} & Gravitational search & Rashedi et al. & \cite{rashedi2009gsa}\\
    Anarchic society optimization & Shayeghi and Dadashpour & \cite{shayeghi2012anarchic} & Harmony search & Geem et al. & \cite{geem2001new}\\
   Artificial cooperative search & Civicioglu & \cite{Civicioglu201358}  & Intelligent water drop & Shah-Hosseini & \cite{shah_hosseini2007problem}\\
   Backtracking optimization search & Civicioglu & \cite{civicioglu2013backtracking} & River formation dynamics & Rabanal et al. & \cite{rabanal2007using}\\
   Differential search algorithm & Civicioglu & \cite{civicioglu2012transforming} & Self-propelled particles & Vicsek & \cite{vicsek1995novel}\\
    Grammatical evolution & Ryan et al. & \cite{ryan1998grammatical} & Simulated annealing & Kirkpatrick et al. & \cite{kirkpatrick1983optimization}\\
  Imperialist competitive algorithm & Atashpaz-Gargari and Lucas & \cite{atashpaz2007imperialist} & Stochastic difusion search & Bishop & \cite{bishop1989stochastic}\\
   League championship algorithm & Kashan & \cite{kashan2009league} & Spiral optimization & Tamura and Yasuda & \cite{tamura2011spiral}\\
   Social emotional optimization & Xu et al. & \cite{xu2010social} & Water cycle algorithm & Eskandar et al. & \cite{eskandar2012water} \\
\hline
\end{tabular}
\caption{A list of algorithms}
\label{Table1}
\end{table}
\end{landscape}
\normalsize

\small
\bibliographystyle{plain}
\bibliography{bibtex}
\normalsize
\vspace{0cm}

\small

\begin{IEEEbiographynophoto}{Iztok Fister Jr.} received his B.Sc. degree in Computer Science in 2011 from the Faculty of Electrical Engineering and Computer Science of the University of Maribor, Slovenia. Currently, he is working towards his M.Sc. degree. His research activities encompasses swarm intelligence, pervasive computing and programming languages.
\end{IEEEbiographynophoto}

\begin{IEEEbiographynophoto}{Xin-She Yang}
received his Ph.D degree in Applied Mathematics from
University of Oxford, UK. Now he is a Reader in Modelling and
Simulation at Middlesex University, UK, an Adjunct Professor at
Reykjavik University, Iceland, and a Distuiguished Guest Professor
at Xi'an Polytechnic University, China. He is also the IEEE CIS
Chair for the Task Force on Business Intelligene and Knowledge
Management, and the Editor-in-Chief of International Journal of
Mathematical Modelling and Numerical Optimisation (IJMMNO).
\end{IEEEbiographynophoto}

\begin{IEEEbiographynophoto}{Iztok Fister} graduated in computer science from the University of
Ljubljana in 1983. He received his Ph.D degree from the Faculty of
Electrical Engineering and Computer Science, University of Maribor,
in 2007. He works as an assistant in the Computer Architecture
and Languages Laboratory at the same faculty. His research interests
include program languages, operational researches and evolutionary
algorithms.
\end{IEEEbiographynophoto}

\begin{IEEEbiographynophoto}{Janez Brest} received his B.S., M.Sc, and Ph.D degrees in computer
science from the University of Maribor, Maribor, Slovenia, in 1995,
1998 and 2000, respectively. He is currently a full professor at the
Faculty of Electrical Engineering and Computer Science, University
of Maribor.
\end{IEEEbiographynophoto}

\begin{IEEEbiographynophoto}{Du\v{s}an Fister} is a student of the Faculty of Electrical Engineering and Computer Science of the University of Maribor. His research activities encompasses GPS solutions, swarm intelligence and operating systems.
\end{IEEEbiographynophoto}

\vfill

\label{finish}
\end{document}